\pdfoutput=1
\documentclass[letterpaper, 10 pt, conference]{ieeeconf}
\IEEEoverridecommandlockouts     

\usepackage{lipsum}

\usepackage{multibib}
\newcites{app}{Appendix References}

\usepackage{wrapfig}
\usepackage{xcolor} 

\usepackage{amsmath}
\usepackage{amsfonts}
\usepackage{amsthm}
\usepackage{amssymb}
\usepackage{mathrsfs}  
\usepackage{bm}
\usepackage[ruled,vlined]{algorithm2e}
\usepackage{mathdots} 
\usepackage{dsfont} 
\usepackage{accents} 
\usepackage{tikz-cd} 
\usepackage{adjustbox} 

\usepackage{caption}
\usepackage{multirow}
\usepackage{multicol}
\usepackage{mdframed} 

\usepackage{listings}
\usepackage{graphicx}

\usepackage{balance}

\newcommand{\comment}[1]{} 
 
\newcommand{\norm}[1]{\left\lVert#1\right\rVert} 
\newcommand\mat[1]{\begin{bmatrix}#1\end{bmatrix}} 
\newcommand\eqs[1]{\begin{equation}\begin{split}#1\end{split}\end{equation}} 
\newcommand\eqsnn[1]{\begin{equation*}\begin{split}#1\end{split}\end{equation*}} 
\DeclareMathOperator*{\setint}{int} 
\DeclareMathOperator*{\dist}{dist}
\DeclareMathOperator*{\diag}{diag} 

\DeclareMathOperator*{\conv}{conv} 
\DeclareMathOperator*{\minimize}{minimize} 
\DeclareMathOperator*{\maximize}{maximize}
\DeclareMathOperator*{\subto}{subject\;to}
\newcommand\parens[1]{\left(#1\right)} 
\newcommand\R{\mathbb{R}} 

\newtheorem{theorem}{Theorem}

\newtheorem*{definition*}{Definition}

\newtheorem{proposition}{Proposition}

\usepackage[bookmarks=true, hidelinks]{hyperref}

\begin{document}

\title{FRoGGeR: \underline{F}ast \underline{Ro}bust \underline{G}rasp \underline{Ge}ne\underline{r}ation via the Min-Weight Metric}

\author{
   Albert H. Li$^\dagger$, Preston Culbertson$^\ddagger$, Joel W. Burdick$^\ddagger$, and Aaron D. Ames$^{\dagger,\ddagger}$%
       \thanks{$\dagger$ A. H. Li and A. D. Ames are with the Department of Computing and Mathematical Sciences, California Institute of Technology, Pasadena, CA 91125, USA, \texttt{\{alberthli, ames\}@caltech.edu}.}%
        \thanks{$\ddagger$ P. Culbertson, J. W. Burdick, and A. D. Ames are with the Department of Civil and Mechanical Engineering, California Institute of Technology, Pasadena, CA 91125, USA, \texttt{\{pculbert, jwb\}@caltech.edu}.}%
}

\maketitle

\begin{abstract}
Many approaches to grasp synthesis optimize analytic quality metrics that measure grasp robustness based on finger placements and local surface geometry. However, generating feasible dexterous grasps by optimizing these metrics is slow, often taking minutes. To address this issue, this paper presents FRoGGeR: a method that quickly generates robust precision grasps using the \textit{min-weight metric}, a novel, almost-everywhere differentiable approximation of the classical $\epsilon$ grasp metric. The min-weight metric is simple and interpretable, provides a reasonable measure of grasp robustness, and admits numerically efficient gradients for smooth optimization. We leverage these properties to rapidly synthesize collision-free robust grasps---typically in less than a second. FRoGGeR can refine the candidate grasps generated by other methods (heuristic, data-driven, etc.) and is compatible with many object representations (SDFs, meshes, etc.). We study FRoGGeR's performance on over 40 objects drawn from the YCB dataset, outperforming a competitive baseline in computation time, feasibility rate of grasp synthesis, and picking success in simulation.  We conclude that FRoGGeR is fast:  it has a median synthesis time of 0.834s over hundreds of experiments. 
\end{abstract}

\maketitle

\section{Introduction}
The success of data-driven methods for grasp synthesis has fundamentally changed manipulation in recent years. Traditional methods \cite{miller2004_graspit} for grasp synthesis focused largely on optimizing analytic quality metrics (e.g., the \textit{largest inscribed ball metric} \cite{kirkpatrick1990_steinitz, ferraricanny1992, rimon2019_manipulationbook}, which we call the \textit{$\epsilon$ metric} as in \cite{kappler2015_bigdatagrasping}), which use hand-object contact points to measure a grasp's robustness to external perturbations. However, these methods suffer some drawbacks in practice: traditional metrics are often hard to optimize and may be non-differentiable, meaning grasps must be synthesized with slower sampling-based methods. Further, they are sensitive to the object geometry (i.e., the surface location and normals), which requires detailed object models to be constructed offline, e.g., by using object scanning rigs \cite{downs2022_google, calli2015ycb}.

To address these shortcomings, a number of authors consider data-driven methods for grasp synthesis (for a detailed survey, see \cite{newbury2022_deep}) that seek to learn grasping policies or metrics depending solely on raw sensor data, such as RGB images or point clouds. A wide variety of approaches have emerged, including using supervised learning to train CNNs to estimate grasp quality from images \cite{mahler2017_dexnet2}, identifying class-level keypoints to find features appropriate for manipulation \cite{manuelli2019_kpam}, or learning a generative model conditioned on depth images \cite{mousavian2019_6dof}. These methods, however, have focused nearly exclusively on generating antipodal grasps for parallel-jaw grippers, or power grasps for dexterous hands. 

In this work, we consider the problem of quickly refining an initial pose for a dexterous hand into a robust precision grasp for a particular object. Compared to power grasps, precision grasps are more useful for manipulation tasks that require delicate or accurate movements, such as tool use or bin packing. Our goals are twofold. First, we seek to generate these grasps quickly (in seconds rather than minutes for current methods \cite{miller2004_graspit, balasubramanian2010_graspit2, turpin2022_diffgraspcontactrich}) while enforcing kinematic and collision constraints. Second, we seek to balance common trade-offs of grasp synthesis methods in terms of performance, speed, and interpretability.

\begin{figure}
    \centering
    \includegraphics[width=\linewidth]{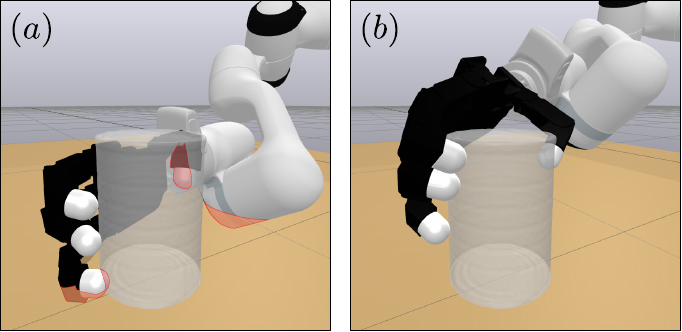}
    \caption{FRoGGeR quickly refines infeasible dexterous grasps into kinematically-feasible, collision-free ones using gradient-based nonlinear optimization. We leverage the \textit{min-weight metric} presented in Sec. \ref{sec:min_weight} to maximize refined grasp robustness. \textbf{(a)} A grasp $q_0$ sampled using a heuristic from Sec. \ref{sec:experiments} with collisions between the object, robot, and table (highlighted red). \textbf{(b)} The feasible, robust refined grasp $q^*$. In this example, the entire process\textemdash from sampling to refinement\textemdash took just 0.816 seconds. The median computation time over all experiments was 0.834 seconds.}
    \label{fig:banner}
    \vspace{-0.5cm}
\end{figure}

\subsection{Contributions}

The main contribution of this work is the formulation of grasp synthesis/refinement as a nonlinear optimization problem that leverages a novel, almost-everywhere differentiable approximation of the $\epsilon$ metric: the \emph{min-weight metric}. The end result is FRoGGeR, a framework for fast robust grasp generation. The optimization problem underlying FRoGGeR can, due to the properties of the min-weight metric, be solved efficiently using commercial solvers. Additionally, FRoGGeR allows us to explicitly enforce constraints on grasps while harnessing the speed of gradient-based optimization.

We aim for our work to be compatible with existing methods in the grasping community. For instance, while FRoGGeR can synthesize grasps with no prior knowledge, it may also refine infeasible or suboptimal grasps computed by learning-based methods. Further, this paper represents object geometry implicitly using \textit{signed distance fields} (SDFs), which means we can leverage existing work that learns SDFs of objects from sensor data \cite{park2019_deepsdf}. To allow the use of mesh-based object representations, we also present practical approximations of the SDF, its gradient, and its Hessian computable using only the mesh.

In summary, our contributions are as follows:
\begin{itemize}
    \item FRoGGeR: a \underline{\textbf{f}}ast \underline{\textbf{ro}}bust \underline{\textbf{g}}rasp \underline{\textbf{ge}}nerato\underline{\textbf{r}} built on the \textit{min-weight metric}, a novel, almost-everywhere differentiable approximation of the $\epsilon$ grasp metric with a numerically efficient gradient;
    \item a practical procedure based on nonlinear optimization with an open-source implementation\footnote{Open-source implementation available at \href{https://github.com/alberthli/frogger}{github.com/alberthli/frogger}.} that generates feasible grasps on the order of seconds; and
    \item numerical experiments and simulations comparing grasps generated by our method to prior work, thereby demonstrating the speed of the proposed approach. 
\end{itemize}

\subsection{Related Work}
The majority of recent work in the grasping literature concerns \textit{parallel-jaw} grasps, which admit simple parameterizations due to the low number of DOFs and ease of control \cite{mahler2017_dexnet2, morrison2018_generativegrasping, eppner2021_acronym, wang2022_dexgraspnet, newbury2022_deep}. Multifinger, dexterous grasping introduces numerous challenges, as the grasp parameterization must specify the states of each finger, and the high dimensionality of this representation demands more fine-grained control and better sensing. Moreover, the complex kinematics and potential for self-collisions complicate the search for feasible grasps. In turn, the problem of synthesizing dexterous grasps, particularly using data-driven methods, has received far less attention than parallel-jaw grasps.

Many classical methods for multifinger grasp synthesis ignore kinematic and collision constraints and only optimize for contact location by leveraging analytic grasp metrics \cite{roa2014_graspmetricssurvey}. The GraspIt! simulator explicitly considers these constraints but simplifies the problem by searching a lower-dimensional space of ``eigengrasps'' using simulated annealing \cite{miller2004_graspit}. This approach has several downsides, including the need to define eigengrasps for new hands (which is highly non-trivial), and slow computational speed in general. Overall, analytic metrics have two main drawbacks: (i) their usage is typically slow, often due to non-smoothness, and (ii) they demand high-fidelity estimates of object geometry and contact locations, diminishing their efficacy \cite{kappler2015_bigdatagrasping}.

In response to these limitations, numerous authors sought to develop data-driven methods for dexterous grasping. Existing approaches include discriminative models, (i.e., those that seek to estimate the quality of a particular grasp), and generative models, which seek to directly generate grasps for novel objects, conditioned on the object geometry or perceptual data. Among these, some only enforce hand kinematics and check collisions post-hoc \cite{kappler2015_bigdatagrasping, aktas2019_deepdexterousgrasping}; others only optimize for contact points and check kinematic feasibility post-hoc \cite{shao2019_unigrasp}; others learn grasp pre-shapes rather than reasoning about contact \cite{lu2020_diffgrasplearning, xu2020_adagrasp}. We refer the reader to \cite{newbury2022_deep} for a detailed survey of data-driven approaches to grasp synthesis. 

One reason for these limitations is the difficulty in casting the nonlinear constrained grasp optimization problem in a computationally tractable way. Among methods addressing this challenge, collision constraints are typically penalized instead of enforced, leading synthesized grasps to have high amounts of infeasible interpenetration \cite{wang2022_dexgraspnet, turpin2022_diffgraspcontactrich, liu2021_diversediffgrasps}. Other attempts at solving the unrelaxed problem are computationally prohibitive (e.g., minutes to hours in \cite{liu2019_graspingmicp}).

In this work, we do not address the perception-based challenges of analytic metrics and assume knowledge of the object's geometry and pose. Instead, we focus on mitigating their slow speed with the view that, despite their drawbacks, these metrics still provide a useful framework for grasping that is agnostic to robot model, object representation, and quality of available data. To that end, we build on prior works that formulate differentiable approximations of analytic force closure measures to recover robust grasps on any multifinger arm/hand system using bilevel optimization.

These prior works propose methods of varying complexity, including solving and differentiating a sequence of linear programs (LPs) \cite{zhu2003_diffgrasplpearly}, a sequence of semidefinite programs (SDPs) \cite{dai2015_forceclosuresdp}, a sum of squares program \cite{liu2020_deepdiffgrasp}, or a single SDP that only approximates force closure \cite{liu2021_diversediffgrasps, wang2022_dexgraspnet}. In contrast, we propose in Sec. \ref{sec:min_weight} a single LP whose optimal value mathematically indicates force closure and whose maximization empirically yields robust grasps. 
%

The method of Wu \textit{et al.} \cite{wu2022_learningdexgraspsgenmodel} is most similar to ours as they also propose solving a bilevel optimization program with smooth collision constraints. However, instead of optimizing for robustness, they solve a feasibility problem and impose a force closure constraint parameterized as a quadratic program while training a conditional variational autoencoder (CVAE) to output performant initial grasps. We compare FRoGGeR's formulation to theirs in Sec. \ref{sec:experiments}.

We do not compare against other analytic metrics in the literature \cite{roa2014_graspmetricssurvey} for two reasons. First, for a fair comparison, they should be differentiable and agnostic to object and task, eliminating many choices such as independent contact regions or task-based methods. Second, among metrics satisfying these desiderata (e.g., the minimum singular value of the grasp matrix), the optimized grasps are not guaranteed to satisfy force closure, so they cannot provide robustness guarantees \textit{even under perfect conditions}.

\subsection{Preliminaries}
Assume a fixed-base, fully-actuated serial manipulator and dexterous hand with $n_c$ fingers contacting the object. Denote by $n$ the total DOFs of the system and $q \in \mathcal{Q} \subset \mathbb{R}^n$ the generalized positions. We let $FK_i(q)$ and $J_i(q)$ denote the forward kinematics and Jacobian of prescribed contact point $i$. Define the hand Jacobian as $J_h=\textrm{blkdiag}(J_1,\dots,J_{n_c})$. We aim to manipulate a rigid object $\mathcal{O}$ with surface $\partial\mathcal{O}$ and body frame $\{O\}$. The pose of a frame $\{F\}$ with respect to a frame $\{G\}$ is expressed $T_{FG} \in SE(3)$. Let $R_{FG} \in SO(3)$ represent the relative position and orientation of $\{G\}$ with respect to $\{F\}$. The robot base frame is denoted $\{B\}$.

We refer to the pair $(q^*,T_{BO})$ as a \textit{grasp}, where $q^*$ is a \textit{feasible} configuration (i.e., no collisions and valid hand-object contact). In this work, we will model the fingers as point contacts with friction. We can thus define $G(q)$, the \textit{grasp map}, which maps a vector of contact forces expressed in their local contact frames, $F_C \in \mathbb{R}^{3 n_c}$, to wrenches in the object frame $w_O \in \mathbb{R}^6$, i.e., we can write $w_O = G(q) F_C$.

We use a Coulomb friction model, so there is no slip if contact forces remain in the \textit{friction cone}, i.e., if $\norm{F^t_C} \leq \mu F^n_C$, where $F^t_C$ and $F^n_C>0$ denote the tangent and normal components respectively. We use a pyramidal friction cone approximation \cite{murray1994_manipulation} with $n_s$ sides and let $m=n_cn_s$ denote the total number of \textit{basis wrenches} forming the finite subset $\mathcal{W}=\{w_i\}_{i=1}^m$ of the \textit{grasp wrench space} $\mathscr{W}\subseteq\R^6$. We let the elements of $\mathcal{W}$ form the columns of the \textit{wrench matrix} $W(q)\in\R^{6\times m}$ and assume there exists a subset of $\mathcal{W}$ containing 7 affinely independent basis wrenches.

We say a grasp is \textit{force closure} if it can resist arbitrary disturbance wrenches in any direction, which is implied if the origin of the grasp wrench space $\mathscr{W}$ lies in the convex hull of $\mathcal{W}$, denoted $\conv(\mathcal{W})$ \cite{ferraricanny1992}. For a thorough treatment of grasping fundamentals, we refer the reader to \cite[Ch.~5]{murray1994_manipulation}.

\section{The Min-Weight Grasp Metric}\label{sec:min_weight}
This section introduces the \textit{min-weight metric}, a simple non-binary indicator of force closure we use as an optimization objective. Specifically, we treat it as a differentiable proxy for the $\epsilon$ metric, which measures the robustness of force closure grasps by reporting the radius of the largest origin-centered ball inscribed in $\conv(\mathcal{W})$ \cite{kirkpatrick1990_steinitz, ferraricanny1992}.

In the sequel, we assume that $\setint(\conv(\mathcal{W}))\neq\emptyset$. To check whether $0\in\conv(\mathcal{W})$, we can solve the following linear feasibility problem \cite{filippozzi2023_convhullmembership} over variables $\alpha = (\alpha_1,\dots,\alpha_m)^\top$, where $\mathds{1}_m\in\R^m$ is the vector of 1s:
\begin{subequations}\label{eqn:feas_prob}
\begin{align}
    \textrm{find} \quad& \alpha \\
    \subto \quad& W\alpha = 0 \label{eqn:convexhullfeasb} \\ 
    & \mathds{1}_m^\top\alpha = 1 \label{eqn:convexhullfeasc} \\ 
    & \alpha \succeq 0. \label{eqn:convexhullfeasd}
\end{align}
\end{subequations}
That is, $0\in\conv(\mathcal{W})\iff\exists \alpha^*$ satisfying \eqref{eqn:convexhullfeasb}-\eqref{eqn:convexhullfeasd}, which is equivalent to the existence of an equilibrium wrench.

\subsection{The Min-Weight Metric \texorpdfstring{$\ell^*(q)$}{} and its Properties}

The key idea of the min-weight metric is to relax constraint \eqref{eqn:convexhullfeasd} by allowing negative weights $\alpha$. If the minimum weight in $\alpha$ is non-negative, the feasibility problem is satisfied, so $0\in\conv(\mathcal{W})$. This motivates the following LP:
\begin{subequations}\label{opt:cvhlp}
\begin{align}
    \ell^*(q) = \maximize_{\alpha\in\R^m,\;\ell\in\R} \quad& \ell \label{eqn:minweighta} \\
    \subto \quad & W(q)\alpha = 0 \label{eqn:minweightb} \\
    & \mathds{1}_m^\top \alpha = 1 \label{eqn:minweightc} \\
    & \alpha \succeq \ell\mathds{1}_m. \label{eqn:minweightd}
\end{align}
\end{subequations}

Thus, $\ell^*(q)$ and $\nabla\ell^*(q)$ are defined even when a grasp is not force closure (i.e., when $\ell^*<0$ in the case of non-physical ``pulling'' contact forces). The dependence of $\ell^*$ on $q$ via the wrench matrix $W(q)$ in constraint \eqref{eqn:minweightb} allows us to iteratively turn suboptimal grasps into force closure ones via gradient-based nonlinear optimization. The following result formally relates $\ell^*$ and force closure status.


\begin{theorem}\label{thm:main}
    If there exists a subset of $\mathcal{W}$ containing 7 affinely independent basis wrenches, then problem \eqref{opt:cvhlp} is feasible. Further, the optimal solution $(\alpha^*,\ell^*)$ satisfies
    \begin{itemize}
        \item [(i)] \textbf{[Non-Force Closure]} $\ell^*(\mathcal{W}) < 0 \iff 0\not\in\conv(\mathcal{W})$,
        \item [(ii)] \textbf{[Robust Closure]} $\ell^*(\mathcal{W}) > 0 \iff 0\in\setint(\conv(\mathcal{W}))$,
        \item [(iii)] \textbf{[Only Equilibrium]} $\ell^*(\mathcal{W}) = 0 \iff 0\in\partial\conv(\mathcal{W})$,
    \end{itemize}
    where $\partial S$ denotes the boundary of a set $S$.
\end{theorem}
\begin{proof}
    For the feasibility claim, it suffices to show that there always exists $\alpha$ satisfying \eqref{eqn:minweightb} and \eqref{eqn:minweightc}, since we can set $\ell$ to its minimum element. Let $W'\in\R^{6\times 7}$ denote a submatrix of $W$ with 7 affinely independent columns and $\alpha'$ the associated weights in $\alpha$. Set all other weights in $\alpha$ to 0. Let $\overline{W}' = \mat{(W')^\top & \mathds{1}_7}^\top\in\R^{7\times 7}$ and $\bar{0}=\mat{0^\top & 1}^\top\in\R^7$.

    The columns of $W'$ are affinely independent in $\R^6$ if and only if the columns of $\overline{W}'$ are linearly independent in $\R^7$ \cite[Exercise 1.1]{brondsted_cvx_polytopes}. Thus, $\overline{W}'$ is invertible, so we can always find $\alpha'$ satisfying $\overline{W}'\alpha'=\bar{0}$. Equivalently, $W'\alpha'=0$ and $\mathds{1}_7^\top\alpha'=1$, implying $W\alpha=0$ and $\mathds{1}_m^\top\alpha=1$.
    
    \textbf{Proof of (i).} By feasibility problem \eqref{eqn:feas_prob}, $\ell^*<0$ implies \eqref{eqn:feas_prob} has no solution, so equivalently, $0\not\in\conv(\mathcal{W})$.
    
    \textbf{Proof of (ii).} $\ell^*(\mathcal{W})>0$ if and only if $\exists\alpha \succ 0$ such that $W\alpha=0$. Since we assume $\conv(\mathcal{W})$ has nonempty interior, its relative interior is its interior, so $0\in\setint(\conv(\mathcal{W}))$ if and only if $\exists\alpha \succ 0$ such that $W\alpha = 0$ \cite[Exercise 3.1]{brondsted_cvx_polytopes}.

    \textbf{Proof of (iii).} Follows immediately from (i) and (ii). \qedhere
\end{proof}

\begin{figure}
    \centering
    \includegraphics[width=\linewidth]{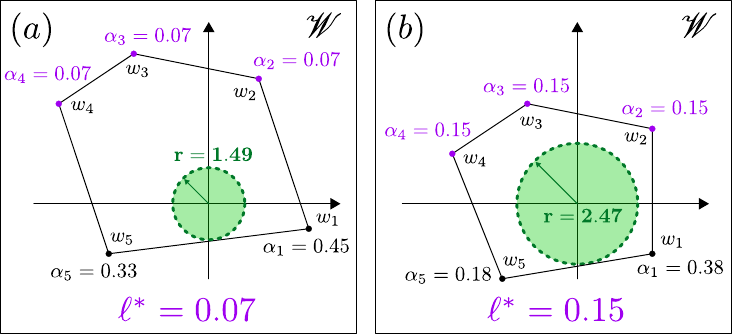}
    \caption{Toy examples of the min-weight metric. The weights $\alpha_i$ are associated with wrenches $w_i \in \mathcal{W} \subseteq \mathscr{W}$. The minimum weights are highlighted purple. The largest inscribed ball about the origin in each convex hull is highlighted green. We empirically observe that the minimum weight $\ell^*$ is strongly correlated with the radius of this ball, i.e., the $\epsilon$ metric.}
    \label{fig:wrench_space}
    \vspace{-0.5cm}
\end{figure}

Thereom \ref{thm:main} states that under mild assumptions, the sign of $\ell^*$ indicates force closure, justifying its maximization. Heuristically, very negative values of $\ell^*$ indicate a grasp is far from force closure while very positive values indicate the origin lies well within $\conv(\mathcal{W})$ (see Fig. \ref{fig:wrench_space}). This motivates using $\ell^*$ as an approximate measure of robustness.

Since $\ell^*\leq 1/m$, the \textit{normalized min-weight metric} $\bar{\ell}^* = m\ell^*$ is well-defined, allowing us to specify the constraint $\bar{\ell}^*\geq k_{\ell}$, where $k_{\ell}\in[0,1]$ is a lower bound on the desired grasp robustness. In experiments, we use $k_\ell=0.3$.

\begin{figure}
    \centering
    \includegraphics[width=\linewidth, trim={0.5cm 0 0.5cm 0.5cm},clip]{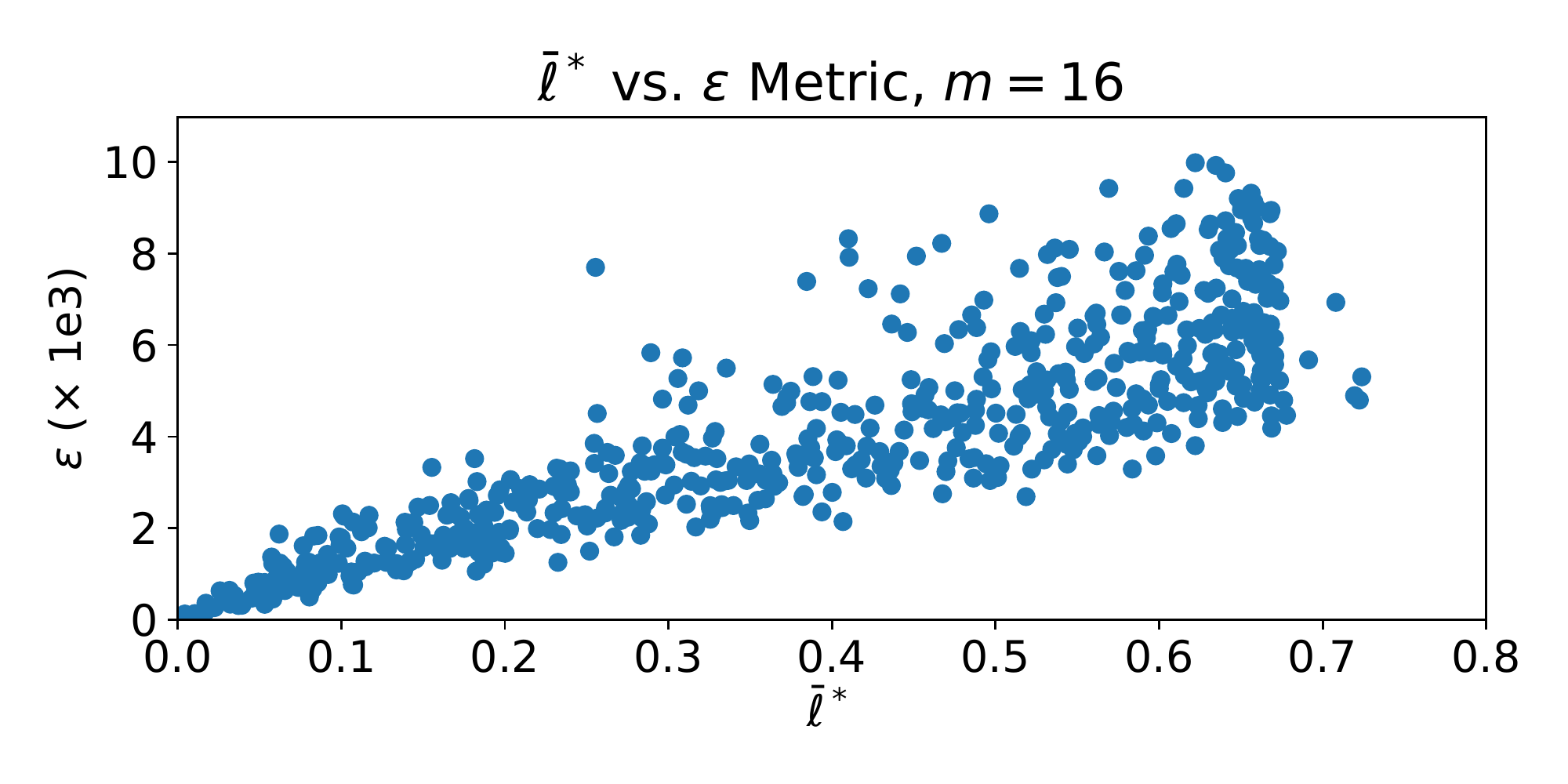}
    \vspace{-1.0cm}
    \caption{$\bar{\ell}^*$ vs. $\epsilon$ metric for 600 total feasible force closure grasps on 6 objects from the YCB dataset \cite{calli2015ycb} generated using FRoGGeR for $m=16$ basis wrenches. We report a Pearson correlation of 0.865 and observe a clear improvement in the worst-case $\epsilon$ value as $\bar{\ell}^*$ increases. Because $\bar{\ell^*}$ is bounded above, we typically observe a ceiling in its value (here, $\approx0.7$) which degrades the correlation as grasp robustness increases. However, we see that $\ell^*$ is very informative in the regime where grasps are barely force closure (bottom left), i.e., where they are least robust.}
    \label{fig:eps_correlation}
    \vspace{-0.5cm}
\end{figure}

We note that while using $\ell^*$ to measure force closure is theoretically justified, its use as a proxy for the $\epsilon$ metric is not, since $\ell^*\gg0$ does not guarantee a large ball is contained in $\conv(\mathcal{W})$. Nevertheless, empirically, we find that $\ell^*$ and the $\epsilon$ metric are strongly correlated and maximizing $\ell^*$ improves a lower bound on the $\epsilon$ value (see Fig. \ref{fig:eps_correlation}). 

Finally, as in many classical metrics, $\ell^*$ is not invariant to the object frame \cite[Ch. 13.5]{rimon2019_manipulationbook}. While methods exist that address this \cite{roa2014_graspmetricssurvey}, we do not explore them in this work.

\subsection{Computing \texorpdfstring{$\nabla \ell^*(q)$}{Dl*(q)} with Differentiable Optimization}
We compute $\nabla\ell^*(q)$ where it is defined using implicit differentiation of the KKT conditions \cite{amos2017_diffopt} and exploit the resulting structure to compute it quickly.

For brevity, let $x=(\alpha,\ell)$ and express \eqref{opt:cvhlp} as
\begin{subequations}
\begin{align}
    \minimize_{x} \quad& c^\top x \\
    \subto \quad & A_{eq}x = b_{eq} \\
    & A_{in}x \preceq 0.
\end{align}
\end{subequations}
Let $\lambda$ and $\nu$ denote the Lagrange multipliers associated with inequality and equality constraints respectively. As in \cite{amos2017_diffopt}, we write the stationarity, primal feasibility, and complementary slackness conditions for \eqref{opt:cvhlp}:
\eqs{\label{eqn:H_expr}
    H = \mat{H_1 \\ H_2 \\ H_3} := \mat{
        c + A_{in}^\top\lambda + A_{eq}^\top\nu \\
        \lambda \odot (A_{in}x) \\
        A_{eq}x - b_{eq}
    },
}
where $\odot$ denotes the Hadamard product. Solving the system $H=0$ is necessary and sufficient to solve any LP since it is convex and always satisfies Slater's condition. Let $D_{(\cdot)}f$ and $\partial_{(\cdot)}f$ denote the total and partial Jacobians of a vector-valued function $f$ with respect to variables $(\cdot)$ respectively. Since $H=0$ at the optimal solution, by implicit differentiation,
\eqs{\label{eqn:kkt_grad_sys}
    &D_{q}H(x^*,\lambda^*,\nu^*,q) = 0 \\
    &\implies \partial_{(x,\lambda,\nu)}H(x^*,\lambda^*,\nu^*,q)D_{q}(x^*,\lambda^*,\nu^*)(q) \\
    &\quad\qquad + \partial_{q}H(x^*,\lambda^*,\nu^*,q) = 0.
}

We can compute $\partial_qH$ explicitly or with autodifferentiation through the primitive wrench matrix $W(q)$, which is derived from the grasp map $G(q)$, with details deferred to our open-source code. We consider the case where $D_{(x,\lambda,\nu)}H$ is invertible and apply the result without further justification like a subgradient in the singular case (similar to the non-differentiable case in \cite{agrawal2019_cvxpylayers}). In particular, $\nabla\ell^*(q)^\top$ is given by the last row of $D_qx^*(q)$, which we can compute via the following result:

\begin{proposition}[Gradient Exploit]\label{prop:gradient_exploit}
    Let $(\cdot)^\dagger$ denote the Moore-Penrose pseudoinverse. Then,
    \eqs{
        D_qx^*(q) = \mat{D_q \alpha^*(q) \\ \nabla \ell^*(q)^\top} = \mat{\diag(\lambda^*)A_{in} \\ A_{eq}}^\dagger \mat{0 \\ \partial_qH_3}.
    }
\end{proposition}

\begin{proof}
    See App. \ref{app:proof}.
\end{proof}

Proposition \ref{prop:gradient_exploit} allows efficient computation of $\nabla\ell^*(q)$, especially when $m$ is ``small'' ($\lesssim40$). For example, when $m=16$ (e.g., using a square pyramidal approximation for a 4-fingered hand), we find that $\ell^*(q)$ and $\nabla\ell^*(q)$ can be computed together in about $3.9ms$ with \verb|cvxpylayers| \cite{agrawal2019_cvxpylayers} on an Nvidia A6000 GPU, whereas using our exploit, they are computed in $0.18ms$ on an Intel i9 CPU, a $22\times$ speedup.

\section{The FRoGGeR Formulation}\label{sec:smooth_opt}
\subsection{The Grasp Refinement Problem}
FRoGGeR \textit{refines} a candidate grasp configuration $q_0$ into a locally optimal one $q^*$ by solving the following nonlinear bilevel optimization program (recalling that $\bar{\ell}^*=m\ell^*$):
\begin{equation}\label{eqn:badly_parameterized_OP}\tag{FRoGGeR}
\begin{split}
    \maximize_q\quad& \ell^*(q) \\
    \subto \quad & q_\text{min} \preceq q \preceq q_\text{max} \\
    &\bar{\ell}^*(q) \geq k_\ell \\
    & FK_i(q) \in \partial\mathcal{O},\; i=1,\dots,n_c \\
    & \text{No (non-finger/object) collision.}
\end{split}
\end{equation}
To enforce joint limits, we constrain the robot configuration $q$ to lie between minimum and maximum values $q_\text{min}$ and $q_\text{max}$. Further, we enforce that the fingertips lie on the object surface $\partial \mathcal{O}$ and that no rigid bodies are interpenetrating. 

To express these constraints mathematically, we first parameterize $\partial\mathcal{O}$ as the 0-level set of a twice-differentiable SDF $s:\R^3\rightarrow\R$, which reports the distance of query points $x\in\R^3$ to $\partial\mathcal{O}$, with $s(x) < 0$ for all points in $\mathcal{O}$:
\begin{equation*}
    s(x) = \begin{cases} -\dist(x, \partial \mathcal{O}), \quad & x \in \mathcal{O},\\
    +\dist(x, \partial \mathcal{O}), & x \notin \mathcal{O}. \end{cases}
\end{equation*}

Second, we consider every possible pair of geometries we would like to prevent from colliding and parameterize the collision status using the differentiable constraints $\sigma\parens{o_A^{(j)},o_B^{(j)};q},\;j=1,\dots,n_p$, where $n_p$ is the number of collision pairs and $\sigma$ is an SDF between two geometries, at least one of whose state depends smoothly on $q$. For pair $j$, we enforce a minimum safety margin of $d_j>0$ unless it is a finger-object pair, for which we specify $d_j<0$ to allow a small amount of interpenetration. Thus, we can express optimization program \eqref{eqn:badly_parameterized_OP} formally as
\begin{subequations}\label{eqn:smooth_opt_program}
\begin{align}
    \maximize\quad& \ell^*(q) \label{eqn:smooth_opt_program_a} \\
    \subto \quad & q_{min} \preceq q \preceq q_{max} \label{eqn:smooth_opt_program_b} \\
    &\bar{\ell}^*(q) \geq k_\ell \label{eqn:smooth_opt_program_c} \\
    & s(FK_i(q))=0,\; i=1,\dots,n_c \label{eqn:smooth_opt_program_d} \\
    & \sigma\parens{o_A^{(j)}, o_B^{(j)};q} \geq d_j,\; j=1,\dots,n_p. \label{eqn:smooth_opt_program_e}
\end{align}
\end{subequations}

\subsection{Gradients of the Constraint Functions}
To use gradient-based methods, we must compute the gradients of the objective and each constraint. The gradient of constraint \eqref{eqn:smooth_opt_program_d} is immediately given by $J_i^\top(q)\nabla s(FK_i(q))$ for $i=1,\dots,n_c$. For constraint \eqref{eqn:smooth_opt_program_e}, we compute for each pair of geometries $(o_A,o_B)$ the \textit{witness points} $(p_A,p_B)$. If the pair is colliding, then the witness points are the two points of furthest penetration. If the pair is not colliding, then they are the two closest points. Let $I_c=1$ indicate collision of a pair and $I_c=0$ otherwise. Then,
\eqs{
    \sigma\parens{o_A, o_B;q} &= (-1)^{I_c+1}\norm{p_A - p_B} \\
    \implies \nabla_q \sigma\parens{o_A, o_B;q} &= (-1)^{I_c}\parens{J_B^\top - J_A^\top}\hat{n}_{AB},
}
where $J_A$ and $J_B$ are the Jacobians at witness points $A$ and $B$ and $\hat{n}_{AB}$ is the unit vector from $p_A$ to $p_B$.

We use \verb|Drake| \cite{drake} to compute witness points for all geometry pairs in a scene. To speed up computation, we represent nonconvex bodies as a union of convex polytopes computed using V-HACD \cite{mamou2009_vhacd}. To reduce the amount of checked pairs, \verb|Drake| culls distant pairs using a broadphase algorithm and we set the associated gradients to 0. When two geometries have exactly 0 signed distance, the gradient may not be defined since $p_A - p_B=0$. In this case, we use the previous value of $\hat{n}_{AB}$, which is initialized randomly.

\begin{figure*}[ht]
    \centering
    \includegraphics[width=0.97647058823\linewidth]{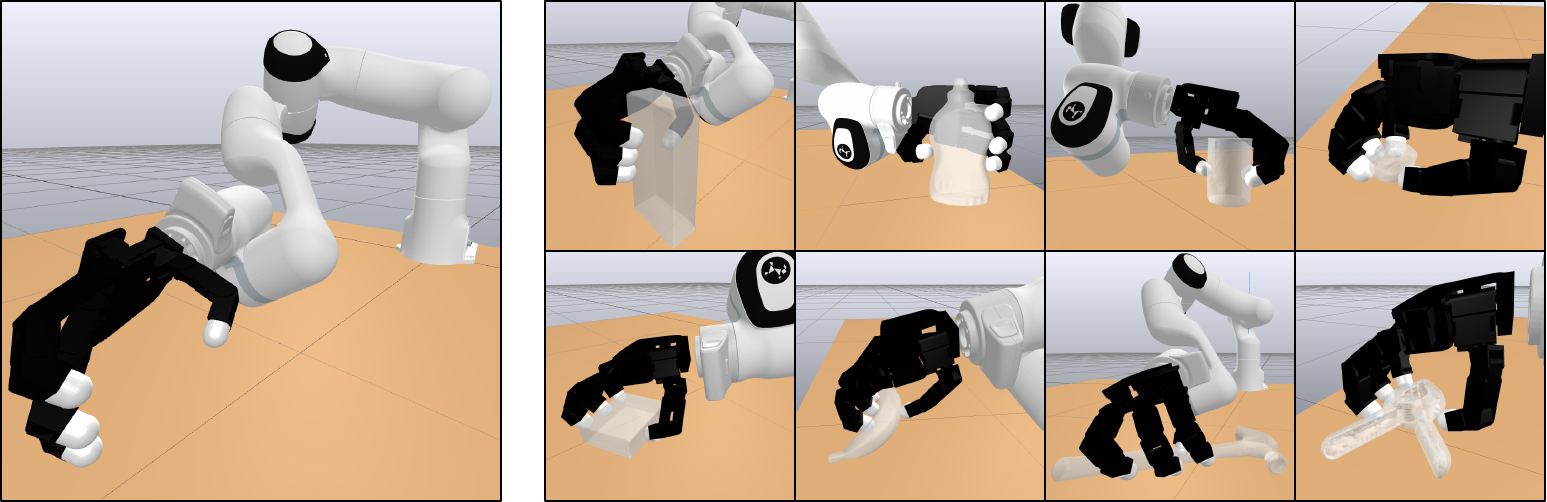}
    \caption{\textbf{Left.} A representative pre-shaped grasp $q_0$ from our heuristic sampler. The initial width of the grip is determined by the object's bounding box. \textbf{Right}. Example refined grasps $q^*$ from our experiments. We can produce robust grasps for highly varying objects that are tall, round, small, flat, long, or otherwise irregular. Top row: sugar box, mustard bottle, soup can, strawberry. Bottom row: pudding box, banana, hammer, large clamp.}
    \label{fig:rep_grasps}
    \vspace{-0.5cm}
\end{figure*}

\subsection{Object Surface Representations}\label{sec:obj_parameterization}
The SDF representation of objects is convenient for reasoning about collision and also geometric properties, since the outward-pointing surface normal at a point $p\in\partial\mathcal{O}$ is given by $\nabla s(p)$ and principal curvatures can be computed from the Hessian $\nabla^2 s(p)$. However, supplying the true object SDF $s$ is non-trivial. Some approaches learn this representation \cite{park2019_deepsdf, merwe2019_learningc3}, while most avoid learning by using the object's mesh or a point cloud \cite{mahler2017_dexnet2, shao2019_unigrasp, liu2020_deepdiffgrasp}. Since these representations are all widely used, it is desirable for grasp synthesis methods to be compatible with any of them.

In the case of a learned or analytical SDF, computing the requisite gradients can be done via autodifferentiation. Further, if provided a dense enough point cloud, the Poisson surface reconstruction algorithm can return a watertight mesh \cite{kazhdan2006_poissonrecon}. Therefore, we focus on the case of meshes.

We assume that there exists a true smooth SDF $s$ and denote the approximation computed with the object mesh as $\tilde{s}$. Then, given a closest point $p'$ to $p$, the gradient is simply
\eqs{\label{eqn:approx_grad}
    \nabla\tilde{s}(p) = \text{sign}(\tilde{s}(p))\frac{p-p'}{\norm{p-p'}},
}
where when $\tilde{s}(p)=0$, $\nabla\tilde{s}(p)$ is the mesh normal at $p'$. To compute $\tilde{s}$ and $\nabla\tilde{s}$, we use the open-source signed distance query provided by \verb|open3d| that also computes the closest point on a mesh to any query point $p\in\R^3$ \cite{zhou2018_open3d}.

To compute whether a grasp is (robustly) force closure, we must compute the grasp map $G(q)$, which depends on the contact frames associated with fingertip positions $p\in\R^3$ \cite{ferraricanny1992}. The normal component of each contact frame is the inward-pointing surface normal, i.e., $\hat{n}(p)=-\nabla s(p)$. Therefore, to differentiate any objective or constraint that depends on measures of force closure, we require $\nabla^2s(p)$.

However, when the object is parameterized as a mesh, the surface is piecewise flat, so $\nabla^2\tilde{s}(p)\equiv0$ wherever it is defined even if $\nabla^2s(p)\neq0$. Other works present methods of varying complexity to compute $\tilde{s}$ and $\nabla \tilde{s}$ that involve solving a quadratic program or deep learning, but do not consider the problem of computing the Hessian of $s$ \cite{liu2020_deepdiffgrasp} \cite{wang2022_dexgraspnet}.

Here, we propose a coarse but efficient approximation. Let $p_1=p\in\R^3$ be an arbitrary query point. Fix a small constant $\delta>0$, randomly select 2 unit vectors denoted $d_2, d_3\in\R^3$, and define $p_i=p_1+\delta d_i$ for $i=2,3$. By \eqref{eqn:approx_grad}, we have $\nabla\tilde{s}(p_i)$ for all $i=1,2,3$. Finally, fix $d_1=\nabla s(p_1)$.

The directional derivative of $\nabla s(p)$ in a direction $v$ is
\eqs{
    \nabla^2 s(p)[v] = \lim_{\delta\rightarrow0}\parens{\nabla s(p+\delta v) - \nabla s(p)}/\delta.
}
Further, for twice-differentiable functions, we must have $\nabla^2 s(p)[v] = \nabla^2 s(p) v$. Thus, using our perturbation directions $d_i$ and a finite-difference approximation of the directional derivatives, $y_i = (\nabla s(p_1 + \delta d_i) - \nabla s(p_1) )/ \delta$, we can write a system of equations to estimate $\nabla^2 s(p)$,
\eqs{
    \Sigma\mat{d_1&d_2&d_3} = \mat{y_1 & y_2 & y_3},
}
by solving for $\Sigma\in\R^{3\times 3}$. We note $y_1=0$ since $\nabla s(p)$ is unchanging (close to the surface) along $\nabla s(p)$ and $d_1,d_2,d_3$ are linearly independent with probability 1. $\Sigma$ denotes an initial estimate for $\nabla^2s(p)$ that may not be symmetric, so we simply choose $\nabla^2\tilde{s}(p) = (\Sigma + \Sigma^\top) / 2$.

The quality of our estimate $\nabla^2\tilde{s}(p)$ is sensitive to the choice of $\delta$ and the properties of the mesh. Empirically, we find that setting $\delta$ to be roughly 10 times the average mesh edge length yields accurate enough gradients for grasp refinement. Using \verb|open3d|, we find the time to estimate all of $\tilde{s}$, $\nabla\tilde{s}$, and $\nabla^2\tilde{s}$ is on the order of $0.1$ms.

\section{Experiments}\label{sec:experiments}

We describe the high-level experimental setup and defer a detailed discussion to App. \ref{app:ctrl}-\ref{app:params}. We use the 7-DOF Franka Research 3 and 16-DOF 4-fingered Allegro hand. The system is mounted on a flat tabletop and each target object is spawned with a fixed initial pose over all trials for repeatability, since the arm/hand configuration is allowed to vary arbitrarily.

We evaluate the robustness of FRoGGeR by executing 20 ``shaky pickups'' per object in simulation using \verb|Drake|. To do so, we generate a pick trajectory where the end-effector is lifted $10cm$ in $1s$ and then held for $1.5s$. We add sinusoidal perturbations to this trajectory with amplitude $3mm$ and varying frequency in all spatial axes $0.25s$ after the pick begins until the end of the simulation. A pick fails if either (1) the object rotates by more than $30^\circ$ or if the object deviates from the pick trajectory by more than $7.5cm$ at any point; or (2) the total grasp synthesis time exceeds 1 minute.

We remark that our shaking test is more dynamic than others in the literature (e.g. \cite{eppner2021_acronym, wu2022_learningdexgraspsgenmodel}), which either do not shake or classify a shake only as a linear movement in space with zero gravity. In contrast, we simulate gravity as well as sustained high-frequency perturbations in all directions.

We compare FRoGGeR's performance on the pickup task to a baseline presented by Wu \textit{et al.}, which only enforces force closure without optimizing for robustness \cite{wu2022_learningdexgraspsgenmodel}.

The controller used in simulation is given by $\tau=J_{h}^\top R_{BC}F_C^* + (I-J_h^\top(J_h^\top)^\dagger)\tau_\text{joint}$. $F_C^*$ is computed by solving for the optimal contact forces to resist external wrenches and errors in the object's pose (e.g., \cite{wu2022_learningdexgraspsgenmodel}). $\tau_\text{joint}$ is the concatenation of arm torques tracking the pick trajectory with hand torques that drive the hand configuration $q_h$ towards the optimized one $q_h^*$. We project $\tau_\text{joint}$ to the null space of $J_h$ to avoid affecting the fingertip locations.

\begin{table*}[t]
    \centering
    \setlength\tabcolsep{2.5pt}
    \begin{tabular}{|c|c||c|c||c|c||c|c|c|}
    \hline
    category & method & \% converged $\uparrow$ & \% pick success $\uparrow$ & $\epsilon$ ($\times$1e3) $\uparrow$ & normalized $\ell^*$ $\uparrow$ & time per solve (s) $\downarrow$ & num. solves $\downarrow$ & total time (s) $\downarrow$ \\
    \hline
    \hline
    sphere & baseline & 12.9\% (31/240) & 83.9\% (26/31) & 2.7 (1.6, 3.6) & 0.32 (0.19, 0.39) & 0.67 (0.44, 1.00) & 68 (61, 73) & 27.2 (13.5, 36.2) \\
    (240 total) & FRoGGeR & \textbf{97.9\%} (235/240) & \textbf{95.3\%} (224/235) & \textbf{4.9} (4.2, 5.5) & \textbf{0.67} (0.55, 0.72) & \textbf{0.31} (0.18, 0.49) & \textbf{2} (1, 4) & \textbf{0.57} (0.30, 1.2) \\
    \hline
    box/cyl & baseline & 52.8\% (169/320) & 68.6\% (116/169) & 2.2 (0.1, 3.8) & 0.20 (0.09, 0.36) & 0.79 (0.41, 1.29) & 42 (13, 53) & 13.4 (5.5, 31.1) \\
    (320 total) & FRoGGeR & \textbf{100\%} (320/320) & \textbf{81.6\%} (261/320) & \textbf{4.8} (3.8, 5.9) & \textbf{0.58} (0.44, 0.65) & \textbf{0.15} (0.09, 0.24) & \textbf{3} (2, 5) & \textbf{0.87} (0.44, 1.6) \\
    \hline
    adversarial & baseline &  61.7\% (185/300) & 43.8\% (81/185) & 1.8 (0.6, 2.9) & 0.18 (0.08, 0.29) & 0.79 (0.47, 1.25) & 29 (10, 55) & 12.7 (5.8, 24.5) \\
    (300 total) & FRoGGeR & \textbf{100\%} (300/300) & \textbf{63.0\%} (189/300) & \textbf{4.3} (3.5, 5.1) & \textbf{0.53} (0.42, 0.63) & \textbf{0.18} (0.11, 0.30) & \textbf{3} (2, 9) & \textbf{1.0} (0.49, 3.3) \\
    \hline
    \hline
    overall & baseline & 44.8\% (385/860) & 58.0\% (223/385) & 2.0 (0.8, 3.4) & 0.19 (0.09, 0.32) & 0.73 (0.44, 1.15) & 50 (17, 63) & 13.8 (5.8, 29.5) \\
    (860 total) & FRoGGeR & \textbf{99.4\%} (855/860) & \textbf{78.8\%} (674/855) & \textbf{4.6} (3.7, 5.5) & \textbf{0.58} (0.45, 0.66) & \textbf{0.21} (0.12, 0.36) & \textbf{3} (1, 6) & \textbf{0.83} (0.39, 1.9) \\
    \hline
    \end{tabular}
    \caption{\textbf{Simulation results.} For each of 43 objects, we try to generate 20 feasible grasps and evaluate them for both methods. In each cell, the top entry is the baseline and the bottom is FRoGGeR. The better result is bolded. $\uparrow$ and $\downarrow$ denote whether higher or lower values are better. Statistics are reported as the median and interquartile range over converged runs and all times are reported in seconds. A run \textit{converges} if it yields a feasible grasp in under 1 minute. We find that our method almost always quickly converges to a feasible grasp while the baseline succeeds under 45\% of the time. Further, over converged grasps, our method outperforms the baseline on a ``shaky pickup'' task in every category and by 20 percentage points overall. Our method's superior robustness is reflected in its $\epsilon$ values, which are about twice as high as the baseline. FRoGGeR's fast convergence is a result of both solving each optimization problem faster as well as requiring significantly fewer attempts before finding a feasible grasp. Note that the total time reported also includes time spent solving IK problems when sampling initial configurations $q_0$ (unreported, since the IK problem solved is the same for both methods).}
    \label{table:results}
    \vspace{-0.5cm}
\end{table*}

To obtain initial configurations $q_0$, we use a heuristic sampler that noisily aligns the palm with the axes of the object's oriented bounding box with probabilities proportional to the box side lengths, motivated by observations of preferred human grasps \cite{balasubramanian2010_inertiagrasp}. We choose a width for the fingertips by computing the width of the appropriate axis of the bounding box. The palm is then placed $4cm$ from the object. To obtain the configuration variables, we solve an inverse kinematics (IK) problem as in \cite{wu2022_learningdexgraspsgenmodel}, but we do not enforce collision constraints or force the fingertips to lie on the object surface. Thus, we only consider infeasible candidate grasps. To solve \eqref{eqn:smooth_opt_program}, we use the \verb|NLopt| \cite{johnson2011_nlopt} implementation of SLSQP \cite{kraft1988_slsqp}.

Our choice to use a coarse sampling heuristic instead of a more performant method is intentional, as our goal is to evaluate FRoGGeR's robustness to the quality of the initial guess. We control for the resulting decrease in performance by evaluating the relative performance of our method versus the baseline under these conditions.

Thus, we do not evaluate the CVAE sampler from \cite{wu2022_learningdexgraspsgenmodel}. Further, we found that the quality of CVAE-generated grasps was not consistent for all objects in our dataset and its performance was on par with our heuristic on a small set of test objects. Ultimately, we chose to synthesize 4-finger grasps to capture the full dexterity of the Allegro hand, which are incompatible with the 3-finger CVAE sampler.

The only difference between our method and the baseline is that in \eqref{eqn:smooth_opt_program}, FRoGGeR maximizes $\ell^*(q)$ with constraint $\bar{\ell}^*(q)\geq 0.3$, while the baseline has no objective and \eqref{eqn:smooth_opt_program_c} is replaced with the bilevel force closure equality constraint described in \cite{wu2022_learningdexgraspsgenmodel}. Otherwise, the same IK routines, sampler, collision geometries, and controller were used.

\subsection{Object Data Processing}
We only present results on objects parameterized as meshes. When supplied with analytical SDFs or well-trained deep SDFs, our method was generally both fast and performant. We use meshes to demonstrate our approach on non-smooth object representations and to validate the usefulness of the Hessian approximation from Sec. \ref{sec:obj_parameterization}.

The objects used in our experiments are from a pruned subset of the YCB dataset \cite{calli2015ycb}. First, we removed all objects that were too large, small, or thin to reasonably grasp with 4 fingers from a flat table, as well as deformable or multibody objects. Second, since the YCB meshes are not watertight, we attempted to reprocess them by densely sampling points on each mesh and running Poisson reconstruction. Of these, we removed objects for which we could not produce watertight meshes due to poor data quality (e.g. from transparency, thin walls, etc.). We note that FRoGGeR works even on non-watertight meshes of adequate quality, but we take this step to eliminate the effect of poor meshes on our results. In total, we test on 43 objects belonging to three categories: \textit{spheroids}, like fruits and balls; \textit{boxes/cylinders}, like food containers, cans, or large cups; and \textit{adversarial} objects with irregular geometry, like tools or very flat/long objects.

For simplicity, we set the friction coefficient to be $\mu=0.7$ for all objects, which is reasonable for the rubbery Allegro fingertips on mostly plastic objects. We also assumed a uniform density of 150 $kg/m^3$ (as in \cite{eppner2021_acronym}) and computed masses using the volume of the processed meshes. The optimizer assumed a more conservative friction coefficient of $\mu=0.5$ and the controller was provided the mass.

\subsection{Results and Discussion}

We report values related to the quality of the grasp (pick success, $\epsilon$ metric value, and $\bar{\ell}^*$) as well as values regarding the runtime of each method in Table \ref{table:results}. We find that overall, FRoGGeR outperforms the baseline in terms of pick success by 20 percentage points and yields $\epsilon$ values that are roughly twice as high. We find that $\bar{\ell}^*$ is a noisy predictor of grasp success - with FRoGGeR, the median and IQR of $\bar{\ell}^*$ were $0.61\;(0.49, 0.67)$ for successes and $0.47\;(0.39, 0.60)$ for failures. See App. \ref{app:hist} for a full histogram.

We also found that overall, FRoGGeR was $\sim$$16\times$ faster at generating grasps than the baseline, a result of $\sim$$3\times$ faster single solve times and $\sim$$15\times$ fewer number of solves required to produce a feasible grasp. This is consistent with the observation by Wu \textit{et al.} that their method struggles to converge to feasible solutions when $q_0$ is infeasible, which requires an expensive IK pre-solve \cite{wu2022_learningdexgraspsgenmodel}. In contrast, FRoGGeR retains superior speed and feasibility rate even with a coarse IK procedure, which suggests that our formulation is also robust to poor candidate grasps. In particular, only 5 runs (all on one object) timed out using our method, whereas over half of the runs timed out for the baseline.

One explanation for this gap is that the baseline force closure equality constraint's gradient vanishes at force closure, which yields a constraint geometry that is difficult to satisfy. Since we do not demand that $q_0$ satisfies \eqref{eqn:smooth_opt_program_d}, we observed that the optimization often terminated unsuccessfully satisfying only one of the equality constraints. In contrast, our constraint \eqref{eqn:smooth_opt_program_c} has non-zero gradients even in force closure, which we conjecture is better-posed numerically, and in particular, allows FRoGGeR to converge for a larger set of candidate grasps $q_0$ than the baseline.

We remark that our reported baseline pick success values are significantly lower than those reported by Wu \textit{et al.} \cite{wu2022_learningdexgraspsgenmodel}, which we attribute to adding shaking to the pick trajectory. When these perturbations were smaller or nonexistent, we typically observed much higher baseline pick success rates, which supports our hypothesis that enforcing only non-robust force closure yields grasps that are brittle in practice.

One of the limitations of our method is that the $\epsilon$ metric often prefers grasps where the fingertips lie on edges or corners, since these regions are typically farther from the center of the object (roughly where we place the object frame), yielding larger moment arms. Moreover, these regions allow a grasp to direct forces in ``non-robust'' directions with little change, drawing solutions to them. This yields unstable grasps in practice, since small deviations in the positions of the fingertips produce large changes in the contact conditions, which commonly occurs in dynamic scenarios.

Edge-seeking behavior was the most common failure mode of both methods, which is reflected by the poor performance on many adversarial objects with less low-curvature area on which to grasp. This behavior was also observed on objects in the box/cyl category, which explains the worse performance compared to spheroids. However, failures often occurred for the baseline even when no fingers were placed on edges.

Finally, we find that the overall performance of both FRoGGeR and the baseline was highly sensitive to the sampled initial conditions. For instance, if the initial width of the fingertips was not guided by object bounding boxes, both methods suffered in terms of runtime and grasp quality, as enforcing surface constraints became harder. This motivates the use of data-driven methods in identifying candidate grasps that may be synergistic with the refinement process.

\section{Conclusion and Future Work}
We presented FRoGGeR, a fast method for generating robust precision grasps using the min-weight metric $\ell^*$, a simple, almost-everywhere differentiable approximation of the $\epsilon$ metric. We have demonstrated that $\ell^*$ is empirically correlated with the $\epsilon$ metric, and validated through simulation that using $\ell^*$ as an optimization objective yields grasps that are more robust to dynamic perturbations than a baseline that only enforces a (non-robust) force closure constraint. Further, we have shown that both the solve time and the feasibility rate of FRoGGeR are superior to that of the baseline.

In the future, we hope to develop methods to combat edge-seeking behavior as well as to generalize FRoGGeR to allow non-precision grasps and a non-fixed number of contact points. Finally, we hope to explore better object representations that do not require online mesh construction or analytical SDFs to be provided beforehand.

\bibliographystyle{unsrt}
\bibliography{references}

\begin{thebibliography}{1}

\bibitem{dewolf2021_lp_subgrad}
Daniel {De Wolf} and Yves Smeers.
\newblock
  \href{https://www.sciencedirect.com/science/article/pii/S0377221719309312}{Generalized
  derivatives of the optimal value of a linear program with respect to matrix
  coefficients}.
\newblock {\em European Journal of Operational Research}, 291(2):491--496,
  2021.

\bibitem{tedrake_manipulationnotes}
Russ Tedrake.
\newblock {\em \href{https://manipulation.mit.edu/index.html}{"Robotic
  Manipulation"}}.
\newblock 2022.

\bibitem{lee2010_se3error}
Taeyoung Lee, Melvin Leok, and N.~Harris McClamroch.
\newblock Geometric tracking control of a quadrotor uav on se(3).
\newblock In {\em \href{https://ieeexplore.ieee.org/document/5717652}{49th IEEE
  Conference on Decision and Control (CDC)}}, pages 5420--5425, 2010.

\bibitem{savva2015_shapenetsem}
M.~Savva, A.~X. Chang, and P.~Hanrahan.
\newblock Semantically-enriched 3d models for common-sense knowledge.
\newblock In {\em
  \href{https://www.computer.org/csdl/proceedings-article/cvprw/2015/07301289/12OmNxFsmtQ}{2015
  IEEE Conference on Computer Vision and Pattern Recognition Workshops
  (CVPRW)}}, pages 24--31, Los Alamitos, CA, USA, jun 2015. IEEE Computer
  Society.

\bibitem{jax2018_github}
James Bradbury, Roy Frostig, Peter Hawkins, Matthew~James Johnson, Chris Leary,
  Dougal Maclaurin, George Necula, Adam Paszke, Jake Vander{P}las, Skye
  Wanderman-{M}ilne, and Qiao Zhang.
\newblock \href{https://jax.readthedocs.io/en/latest/index.html}{{JAX}:
  composable transformations of {P}ython+{N}um{P}y programs}, 2018.

\bibitem{trimesh}
{Dawson-Haggerty et al.}
\newblock \href{https://github.com/mikedh/trimesh}{trimesh}, 2019.

\bibitem{quantecon}
QuantEcon Organization.
\newblock \href{https://github.com/QuantEcon/QuantEcon.py}{The QuantEcon
  package.}, 2023.

\bibitem{lam2015_numba}
Siu~Kwan Lam, Antoine Pitrou, and Stanley Seibert.
\newblock \href{https://numba.pydata.org/}{Numba: A llvm-based python jit
  compiler}.
\newblock In {\em Proceedings of the Second Workshop on the LLVM Compiler
  Infrastructure in HPC}, pages 1--6, 2015.

\bibitem{pan2012_fcl}
Jia Pan, Sachin Chitta, and Dinesh Manocha.
\newblock \href{https://ieeexplore.ieee.org/document/6225337}{FCL: A general
  purpose library for collision and proximity queries}.
\newblock In {\em 2012 IEEE International Conference on Robotics and
  Automation}, pages 3859--3866, 2012.

\end{thebibliography}


\begin{thebibliography}{10}

\bibitem{miller2004_graspit}
A.T. Miller and P.K. Allen.
\newblock \href{https://ieeexplore.ieee.org/document/1371616}{Graspit! A
  versatile simulator for robotic grasping}.
\newblock {\em IEEE Robotics and Automation Magazine}, 11(4):110--122, 2004.

\bibitem{kirkpatrick1990_steinitz}
D.~G. Kirkpatrick, B.~Mishra, and C.~K. Yap.
\newblock Quantitative steinitz's theorems with applications to multifingered
  grasping.
\newblock In {\em Proceedings of the Twenty-Second Annual ACM Symposium on
  Theory of Computing}, STOC '90, page 341–351, New York, NY, USA, 1990.
  Association for Computing Machinery.

\bibitem{ferraricanny1992}
C.~Ferrari and J.~Canny.
\newblock \href{https://ieeexplore.ieee.org/document/219918}{Planning optimal
  grasps}.
\newblock In {\em Proceedings 1992 IEEE International Conference on Robotics
  and Automation}, pages 2290--2295 vol.3, 1992.

\bibitem{rimon2019_manipulationbook}
Elon Rimon and Joel Burdick.
\newblock {\em The Mechanics of Robot Grasping}.
\newblock Cambridge University Press, 2019.

\bibitem{kappler2015_bigdatagrasping}
Daniel Kappler, Jeannette Bohg, and Stefan Schaal.
\newblock \href{https://ieeexplore.ieee.org/document/7139793}{Leveraging big
  data for grasp planning}.
\newblock In {\em 2015 IEEE International Conference on Robotics and Automation
  (ICRA)}, pages 4304--4311, 2015.

\bibitem{downs2022_google}
Laura Downs, Anthony Francis, Nate Koenig, Brandon Kinman, Ryan Hickman, Krista
  Reymann, Thomas~B. McHugh, and Vincent Vanhoucke.
\newblock \href{https://arxiv.org/abs/2204.11918}{Google Scanned Objects: {{A}}
  High-Quality Dataset of {{3D}} Scanned Household Items}.
\newblock In {\em 2022 International Conference on Robotics and Automation
  ({{ICRA}})}, pages 2553--2560, 2022.

\bibitem{calli2015ycb}
Berk Calli, Arjun Singh, Aaron Walsman, Siddhartha Srinivasa, Pieter Abbeel,
  and Aaron~M. Dollar.
\newblock \href{https://ieeexplore.ieee.org/document/7251504}{The {{YCB}}
  Object and {{Model}} Set: {{Towards}} Common Benchmarks for Manipulation
  Research}.
\newblock In {\em 2015 {{International Conference}} on {{Advanced Robotics}}
  ({{ICAR}})}, pages 510--517, July 2015.

\bibitem{newbury2022_deep}
Rhys Newbury, Morris Gu, Lachlan Chumbley, Arsalan Mousavian, Clemens Eppner,
  J{\"u}rgen Leitner, Jeannette Bohg, Antonio Morales, Tamim Asfour, Danica
  Kragic, Dieter Fox, and Akansel Cosgun.
\newblock \href{https://arxiv.org/abs/2207.02556}{Deep Learning Approaches to
  Grasp Synthesis: {{A}} Review}, 2022.

\bibitem{mahler2017_dexnet2}
Jeffrey Mahler, Jacky Liang, Sherdil Niyaz, Michael Laskey, Richard Doan, Xinyu
  Liu, Juan~Aparicio Ojea, and Ken Goldberg.
\newblock \href{https://arxiv.org/abs/1703.09312}{Dex-Net 2.0: Deep Learning to
  Plan Robust Grasps with Synthetic Point Clouds and Analytic Grasp Metrics}.
\newblock {\em ArXiv}, abs/1703.09312, 2017.

\bibitem{manuelli2019_kpam}
Lucas Manuelli, Wei Gao, Peter Florence, and Russ Tedrake.
\newblock \href{https://arxiv.org/abs/1903.06684}{{{kPAM}}: {{KeyPoint
  Affordances}} for {{Category-Level Robotic Manipulation}}}.
\newblock {\em arXiv:1903.06684 [cs]}, October 2019.

\bibitem{mousavian2019_6dof}
Arsalan Mousavian, Clemens Eppner, and Dieter Fox.
\newblock
  \href{https://openaccess.thecvf.com/content_ICCV_2019/papers/Mousavian_6-DOF_GraspNet_Variational_Grasp_Generation_for_Object_Manipulation_ICCV_2019_paper.pdf}{6-{{DOF
  GraspNet}}: {{Variational Grasp Generation}} for {{Object Manipulation}}}.
\newblock In {\em 2019 {{IEEE}}/{{CVF International Conference}} on {{Computer
  Vision}} ({{ICCV}})}, pages 2901--2910, {Seoul, Korea (South)}, October 2019.
  {IEEE}.

\bibitem{balasubramanian2010_graspit2}
Ravi Balasubramanian, Ling Xu, Peter~D. Brook, Joshua~R. Smith, and Yoky
  Matsuoka.
\newblock \href{https://ieeexplore.ieee.org/document/5509855}{Human-guided
  grasp measures improve grasp robustness on physical robot}.
\newblock In {\em 2010 IEEE International Conference on Robotics and
  Automation}, pages 2294--2301, 2010.

\bibitem{turpin2022_diffgraspcontactrich}
Dylan Turpin, Liquan Wang, Eric Heiden, Yun-Chun Chen, Miles Macklin, Stavros
  Tsogkas, Sven Dickinson, and Animesh Garg.
\newblock \href{https://arxiv.org/abs/2208.12250}{Grasp’D: Differentiable
  Contact-Rich Grasp Synthesis for Multi-Fingered Hands}.
\newblock In {\em Computer Vision – ECCV 2022: 17th European Conference, Tel
  Aviv, Israel, October 23–27, 2022, Proceedings, Part VI}, page 201–221,
  2022.

\bibitem{park2019_deepsdf}
Jeong~Joon Park, Peter~R. Florence, Julian Straub, Richard~A. Newcombe, and
  S.~Lovegrove.
\newblock \href{https://arxiv.org/abs/1901.05103}{DeepSDF: Learning Continuous
  Signed Distance Functions for Shape Representation}.
\newblock {\em 2019 IEEE/CVF Conference on Computer Vision and Pattern
  Recognition (CVPR)}, pages 165--174, 2019.

\bibitem{morrison2018_generativegrasping}
Douglas Morrison, Peter Corke, and J.~Leitner.
\newblock \href{https://arxiv.org/abs/1804.05172}{Closing the Loop for Robotic
  Grasping: A Real-time, Generative Grasp Synthesis Approach}{Closing the Loop
  for Robotic Grasping: A Real-time, Generative Grasp Synthesis Approach}.
\newblock {\em ArXiv}, abs/1804.05172, 2018.

\bibitem{eppner2021_acronym}
Clemens Eppner, Arsalan Mousavian, and Dieter Fox.
\newblock \href{https://arxiv.org/abs/2011.09584}{ACRONYM: A Large-Scale Grasp
  Dataset Based on Simulation}.
\newblock In {\em 2021 IEEE International Conference on Robotics and Automation
  (ICRA)}, pages 6222--6227, 2021.

\bibitem{wang2022_dexgraspnet}
Ruicheng Wang, Jialiang Zhang, Jiayi Chen, Yinzhen Xu, Puhao Li, Tengyu Liu,
  and He~Wang.
\newblock \href{https://arxiv.org/abs/2210.02697}{DexGraspNet: A Large-Scale
  Robotic Dexterous Grasp Dataset for General Objects Based on Simulation}.
\newblock {\em ArXiv}, abs/2210.02697, 2022.

\bibitem{roa2014_graspmetricssurvey}
M{\'a}ximo~A. Roa and Ra{\'u}l Su{\'a}rez.
\newblock
  \href{https://link.springer.com/article/10.1007/s10514-014-9402-3}{Grasp
  quality measures: review and performance}.
\newblock {\em Autonomous Robots}, 38:65 -- 88, 2014.

\bibitem{aktas2019_deepdexterousgrasping}
Umit~Rusen Aktas, Chaoyi Zhao, Marek Kopicki, Ales Leonardis, and Jeremy~L.
  Wyatt.
\newblock \href{https://arxiv.org/abs/1908.04293}{Deep Dexterous Grasping of
  Novel Objects from a Single View}.
\newblock {\em Int. J. Humanoid Robotics}, 19:2250011:1--2250011:30, 2019.

\bibitem{shao2019_unigrasp}
Lin Shao, F{\'a}bio Ferreira, Mikael Jorda, Varun Nambiar, Jianlan Luo, Eugen
  Solowjow, Juan~Aparicio Ojea, Oussama Khatib, and Jeannette Bohg.
\newblock \href{https://arxiv.org/abs/1910.10900}{UniGrasp: Learning a Unified
  Model to Grasp with N-Fingered Robotic Hands}.
\newblock {\em ArXiv}, abs/1910.10900, 2019.

\bibitem{lu2020_diffgrasplearning}
Qingkai Lu, Mark Van~der Merwe, Balakumar Sundaralingam, and Tucker Hermans.
\newblock \href{https://arxiv.org/abs/2001.09242}{Multifingered Grasp Planning
  via Inference in Deep Neural Networks: Outperforming Sampling by Learning
  Differentiable Models}.
\newblock {\em IEEE Robotics and Automation Magazine}, 27(2):55--65, 2020.

\bibitem{xu2020_adagrasp}
Zhenjia Xu, Beichun Qi, Shubham Agrawal, and Shuran Song.
\newblock \href{https://arxiv.org/abs/2011.14206}{AdaGrasp: Learning an
  Adaptive Gripper-Aware Grasping Policy}.
\newblock {\em 2021 IEEE International Conference on Robotics and Automation
  (ICRA)}, pages 4620--4626, 2020.

\bibitem{liu2021_diversediffgrasps}
Tengyu Liu, Zeyu Liu, Ziyuan Jiao, Yixin Zhu, and Song-Chun Zhu.
\newblock \href{https://arxiv.org/abs/2104.09194}{Synthesizing Diverse and
  Physically Stable Grasps With Arbitrary Hand Structures Using Differentiable
  Force Closure Estimator}.
\newblock {\em IEEE Robotics and Automation Letters}, 7:470--477, 2021.

\bibitem{liu2019_graspingmicp}
Min Liu, Zherong Pan, Kai Xu, and Dinesh Manocha.
\newblock \href{https://arxiv.org/abs/1909.05430}{New Formulation of
  Mixed-Integer Conic Programming for Globally Optimal Grasp Planning}.
\newblock {\em IEEE Robotics and Automation Letters}, 5:4663--4670, 2019.

\bibitem{zhu2003_diffgrasplpearly}
Xiangyang Zhu and Jun Wang.
\newblock
  \href{https://ieeexplore.ieee.org/abstract/document/1220716}{Synthesis of
  force-closure grasps on 3-D objects based on the Q distance}.
\newblock {\em IEEE Trans. Robotics Autom.}, 19:669--679, 2003.

\bibitem{dai2015_forceclosuresdp}
Hongkai Dai, Anirudha Majumdar, and Russ Tedrake.
\newblock
  \href{https://link.springer.com/chapter/10.1007/978-3-319-51532-8_18}{Synthesis
  and Optimization of Force Closure Grasps via Sequential Semidefinite
  Programming}.
\newblock In {\em International Symposium of Robotics Research}, 2015.

\bibitem{liu2020_deepdiffgrasp}
Min Liu, Zherong Pan, Kai Xu, Kanishka Ganguly, and Dinesh Manocha.
\newblock \href{https://arxiv.org/abs/2002.01530}{Deep Differentiable Grasp
  Planner for High-DOF Grippers}.
\newblock {\em ArXiv}, abs/2002.01530, 2020.

\bibitem{wu2022_learningdexgraspsgenmodel}
Albert Wu, Michelle Guo, and C.~Karen Liu.
\newblock \href{https://arxiv.org/abs/2207.00195}{Learning Diverse and
  Physically Feasible Dexterous Grasps with Generative Model and Bilevel
  Optimization}.
\newblock {\em ArXiv}, abs/2207.00195, 2022.

\bibitem{murray1994_manipulation}
Richard~M Murray, Zexiang Li, and S~Shankar Sastry.
\newblock {\em A mathematical introduction to robotic manipulation}.
\newblock CRC press, 1994.

\bibitem{filippozzi2023_convhullmembership}
Rafaela Filippozzi, Douglas~S. Gonçalves, and Luiz-Rafael Santos.
\newblock
  \href{https://www.sciencedirect.com/science/article/pii/S037722172200683X}{First-order
  methods for the convex hull membership problem}.
\newblock {\em European Journal of Operational Research}, 306(1):17--33, 2023.

\bibitem{brondsted_cvx_polytopes}
Arne Brøndsted.
\newblock {\em An Introduction to Convex Polytopes}.
\newblock Graduate Texts in Mathematics. Springer-Verlag, New York, 1982.

\bibitem{amos2017_diffopt}
Brandon Amos and J.~Zico Kolter.
\newblock \href{https://arxiv.org/abs/1703.00443}{OptNet: Differentiable
  Optimization as a Layer in Neural Networks}.
\newblock In {\em International Conference on Machine Learning}, 2017.

\bibitem{agrawal2019_cvxpylayers}
A.~Agrawal, B.~Amos, S.~Barratt, S.~Boyd, S.~Diamond, and Z.~Kolter.
\newblock
  \href{https://papers.nips.cc/paper/2019/hash/9ce3c52fc54362e22053399d3181c638-Abstract.html}{Differentiable
  Convex Optimization Layers}.
\newblock In {\em Advances in Neural Information Processing Systems}, 2019.

\bibitem{drake}
Russ Tedrake and the Drake Development~Team.
\newblock \href{https://drake.mit.edu}{Drake: Model-based design and
  verification for robotics}, 2019.

\bibitem{mamou2009_vhacd}
Khaled Mamou and Faouzi Ghorbel.
\newblock \href{https://ieeexplore.ieee.org/document/5414068}{A simple and
  efficient approach for 3D mesh approximate convex decomposition}.
\newblock In {\em 2009 16th IEEE International Conference on Image Processing
  (ICIP)}, pages 3501--3504, 2009.

\bibitem{merwe2019_learningc3}
Mark~Van der Merwe, Qingkai Lu, Balakumar Sundaralingam, Martin Matak, and
  Tucker Hermans.
\newblock \href{https://arxiv.org/abs/1910.00983}{Learning Continuous 3D
  Reconstructions for Geometrically Aware Grasping}.
\newblock {\em 2020 IEEE International Conference on Robotics and Automation
  (ICRA)}, pages 11516--11522, 2019.

\bibitem{kazhdan2006_poissonrecon}
Michael Kazhdan, Matthew Bolitho, and Hugues Hoppe.
\newblock \href{https://diglib.eg.org/handle/10.2312/SGP.SGP06.061-070}{Poisson
  Surface Reconstruction}.
\newblock In Alla Sheffer and Konrad Polthier, editors, {\em Symposium on
  Geometry Processing}. The Eurographics Association, 2006.

\bibitem{zhou2018_open3d}
Qian-Yi Zhou, Jaesik Park, and Vladlen Koltun.
\newblock \href{https://arxiv.org/abs/1801.09847}{{Open3D}: {A} Modern Library
  for {3D} Data Processing}.
\newblock {\em arXiv:1801.09847}, 2018.

\bibitem{balasubramanian2010_inertiagrasp}
Ravi Balasubramanian, Ling Xu, Peter~D. Brook, Joshua~R. Smith, and Yoky
  Matsuoka.
\newblock \href{https://ieeexplore.ieee.org/document/5509855}{Human-guided
  grasp measures improve grasp robustness on physical robot}.
\newblock In {\em 2010 IEEE International Conference on Robotics and
  Automation}, pages 2294--2301, 2010.

\bibitem{johnson2011_nlopt}
Steven~G. Johnson.
\newblock \href{http://ab-initio.mit.edu/nlopt}{The NLopt
  nonlinear-optimization package}, 2011.

\bibitem{kraft1988_slsqp}
Dieter Kraft.
\newblock
  \href{http://degenerateconic.com/uploads/2018/03/DFVLR_FB_88_28.pdf}{A
  software package for sequential quadratic programming}.
\newblock {\em Forschungsbericht Deutsche Forschungs und Versuchsanstalt fur
  Luft und Raumfahrt}, 1988.

\end{thebibliography}

\newpage

\appendix
This appendix provides the proof of Proposition \ref{prop:gradient_exploit} as well as numerous implementation details. For the most fine-grained explanation of these details, we refer the reader to our open-source implementation at \href{https://github.com/alberthli/frogger}{github.com/alberthli/frogger}.

\subsection{Proof of Proposition \ref{prop:gradient_exploit}}\label{app:proof}
By direct computation, we have
\eqs{
    &\partial_{(x,\lambda,\nu)}H = \mat{
        0 & A_{in}^\top & A_{eq}^\top \\
        \diag(\lambda^*)A_{in} & \diag(A_{in}x^*) & 0 \\
        A_{eq} & 0 & 0
    }.
}
For brevity, let $\Omega:=\partial_{(x,\lambda,\nu)}H$ and unless otherwise stated, let functions be evaluated at the optimal primal/dual solution $(x^*,\lambda^*,\nu^*)$. Let
\eqs{
    \Omega := \mat{A & B \\ C & D},
}
where for convenience we denote
\eqsnn{
    A = 0_{(m+1)\times(m+1)}&,\quad B = \mat{A_{in}^\top & A_{eq}^\top}, \\
    C = \mat{\diag(\lambda^*)A_{in} \\ A_{eq}}&,\quad D = \mat{\diag(A_{in}x^*)&0_{m\times7}\\0_{7\times m} & 0_{7\times7}}.
}
\begin{proof}
    We have
    \eqs{
        \Omega^{-1} &= (\Omega^\top\Omega)^{-1}\Omega^\top \\
        &= \mat{C^\top C & C^\top D \\ D^\top C & B^\top B + D^\top D}^{-1}\mat{0 & C^\top \\ B^\top & D^\top}.
    }
    Observe that
    \eqs{
        C^\top D &= \mat{A^\top_{in}\diag(\lambda^*) & A_{eq}^\top}\mat{\diag(A_{in}x^*) & 0 \\ 0& 0} \\
        &= \mat{A_{in}^\top\diag(\lambda^*)\diag(A_{in}x^*) & 0_{(m+1)\times 7}} \\
        &= 0,
    }
    where the last equality follows because
    \eqs{
        \diag(\lambda^*)\diag(A_{in}x^*)=\diag(\lambda^*\odot(A_{in}x^*))=0
    }
    by complementary slackness. Letting $P=C^\top C$ and $R = B^\top B + D^\top D$,
    \eqs{\label{eqn:omega_inverse}
        \Omega^{-1} &= \mat{P^{-1} & 0 \\ 0 & R^{-1}}\mat{0 & C^\top \\ B^\top & D^\top} \\
        &= \mat{0 & P^{-1}C^\top \\ R^{-1}B^\top & R^{-1}D^\top},
    }
    where we note that $P^{-1}C^\top=C^\dagger$. By substituting \eqref{eqn:omega_inverse} into \eqref{eqn:kkt_grad_sys}, the result immediately follows.
\end{proof}

We remark that if $\ell^*$ is locally Lipschitz with respect to the constraint matrix parameters $A_{eq}$ and $A_{in}$, it is differentiable everywhere but a set of measure 0 by a theorem of Rademacher (see \citeapp{dewolf2021_lp_subgrad} for discussion). Further, the gradient is defined when \eqref{opt:cvhlp} has unique primal/dual optima and in this case, $\parens{\partial_{(x,\lambda,\nu)}H}^{-1}$ is defined so $\nabla\ell^*(q)$ is computable \citeapp[Prop. 4.1]{dewolf2021_lp_subgrad}. The Lipschitz condition can always be satisfied by removing degenerate constraints, so we assume it.

\subsection{Controller Implementation Details}\label{app:ctrl}
This section explains the structure of the controller used for the pickup task. Recall that our controller is of the form
\eqs{\label{eqn:pick_controller}
    \tau=J_{h}^\top R_{BC}F_C^* + (I-J_h^\top(J_h^\top)^\dagger)\tau_\text{joint}.
}

We first explain computation of $\tau_{\text{joint}}\in\R^n$. We have that
\eqs{
    \tau_{\text{joint}} = \tau_{\text{grav}} + \tau_{\text{track}},
}
where the first term is a gravity compensation torque computed using partial inverse dynamics (i.e., ignoring the inertial/coriolis dynamical terms and assuming quasi-static operation) and the second term is a tracking term with independent components for the arm and the hand. We have
\eqs{
    \tau_{\text{track}} = \mat{\tau_{\text{arm}} \\ \tau_{\text{hand}}}.
}

The arm tracking torques are computed as
\eqs{
    \tau_{\text{arm}} = -K_{p,\text{arm}}(q_a-q_{a,\text{des}}) - K_{d,\text{arm}}(\dot{q}_a-\dot{q}_{a,\text{des}}),
}
with the gains set to
\eqs{
    K_{p,\text{arm}} &= 500I, \\
    K_{d,\text{arm}} &= \diag\parens{\mat{1, 1, 1, 1, 0.1, 0.1, 0.1}}.
}
The desired values $q_{a,\text{des}}$ and $\dot{q}_{a,\text{des}}$ are computed via a differential inverse kinematics controller implemented in \verb|Drake| that converts a desired end-effector pose trajectory specified in Cartesian space to joint angles and velocities that can be tracked. We defer those details to \citeapp[Ch. 3.10]{tedrake_manipulationnotes}.

The hand tracking torques are simply given by the following proportional controller:
\eqs{
    \tau_{\text{hand}} = -k_{p,\text{hand}}(q_h - q_h^*),
}
where $q_h^*$ is the component of the refined configuration $q^*$ corresponding to the hand states and $k_{p,\text{hand}}=5$. We project all of these torques via the left multiplication of $(I-J_h^\top(J_h^\top)^\dagger)$ to the null space of $J_h$, which ensures that applying them does not change the contact positions between the hand and object. We note that this projection does not affect the arm torques at all.

We now explain the computation of the optimal contact forces $F_C^*$, which is formulated as the solution to the following quadratic program:
\eqsnn{
    \minimize_{F_C}\quad& \norm{GF_C- \parens{-{}^Ow_{\text{des}}}}_2^2 \\
    \subto \quad & \Lambda_iF_{C,i} \leq 0, \; i = 1,\dots,n_c \\
    & F_{C,i}^n \geq F_{\text{min}}^n, \; i = 1,\dots,n_c \\
    & \tau_{\text{lb}} - \tau_{\text{joint,h}} \leq J_i^\top R_{BC}F_{C,i}, \; i=1,\dots,n_c \\
    & J_i^\top R_{BC}F_{C,i} \leq \tau_{\text{ub}} - \tau_{\text{joint,h}}, \; i=1,\dots,n_c.
}
We note that $F_{C,i}\in\R^3$ is the $i^{th}$ contact force in the concatenation of all contact forces $F_C$.

The QP objective produces applied wrenches on the object as close as possible to counteracting some desired wrench whose force and torque components are expressed in the object frame. The first constraint represents the pyramidal friction cone constraints (e.g., \cite{wu2022_learningdexgraspsgenmodel}). As in \cite{wu2022_learningdexgraspsgenmodel}, the second enforces a minimum normal force which we specify as $F_{\text{min}}^n=1.0$ and we additionally specify that if the object weighs under $0.01kg$, we set $F_{\text{min}}^n=0.25$. The final two constraints enforce torque limits by ensuring the total applied joint torques from both controller terms in \eqref{eqn:pick_controller} respect the desired limits.

The desired external wrench is computed as follows:
\eqs{
    {}^Ow_{\text{des}} = R_{OB}\parens{{}^Bw_{\text{grav}} + {}^Bw_{\text{err}}},
}
where ${}^Bw_{\text{grav}}$ is the gravitational wrench expressed in the robot base frame. ${}^Bw_{\text{err}}$ is an error wrench computed from the measured error in the object's desired pose. Suppose the desired object pose in the world frame is specified as the tuple $(p_{\text{des}},R_{\text{des}})$ where we suppress the frame notation for brevity. We convert errors in the pose into a wrench using the formulation provided in \citeapp{lee2010_se3error}:
\eqs{
    {}^Bw_{\text{err}} = \mat{-k_{p,\text{err}}(p-p_{\text{des}}) - k_{d,\text{err}}\dot{p} \\ -k_{R,\text{err}}e_R - k_{\omega,\text{err}}\omega},
}
where
\eqs{
    e_R &= \parens{\frac{1}{2}\parens{R_{\text{des}}^\top R - R^\top R_{\text{des}}}^\vee},
}
$(\cdot)^\vee$ is the map sending elements of the Lie algebra $so(3)$ to $\R^3$ (see \citeapp{lee2010_se3error}), and $\omega$ is the angular velocity of the object computed using numerical differentiation of its orientation. We choose the gains
\eqs{
    k_{p,\text{err}} &= 50, \\
    k_{d,\text{err}} &= 5, \\
    k_{R,\text{err}} &= 50, \\
    k_{\omega,\text{err}} &= 5.
}

\subsection{Heuristic Sampler Implementation Details}\label{app:sampler}
We first fix a convention for the axes of the palm of the hand. Let the $x$-axis be the outward palm normal and the $z$-axis be the corresponding axis that points in the direction of the fingers of the hand (for non-anthropomorphic hands, this choice may be arbitrary). The $y$-axis is then chosen consistently with the right hand rule.

The heuristic sampler consists of the following steps: (1) from the oriented bounding box of the object (which can be computed approximately very quickly using \verb|open3d|), choose an axis with which to align the palm's $y$-axis up to sign and use the width of this box edge to fix an initial guess for the separation of the hand's fingers; (2) of the two remaining axes, choose one with which to align the palm's $x$-axis; (3) add rotational noise drawn from the von Mises distribution on the 2-sphere to randomly perturb the palm frame; (4) compute a desired location of the palm frame with respect to the object by placing it roughly $4cm$ from the surface of the object, which is approximated as its bounding box; (5) using the constraints on the palm frame, solve an inverse kinematics problem to recover $q_0$.

The probability of choosing a given bounding box axis for alignment is proportional to its length. For instance, if the bounding box has side lengths $a, b, c$, then the probability of choosing the first axis is $a/(a+b+c)$. For very short objects, we only accepted a palm frame whose $x$-axis approached the object from above to avoid heavy collisions with the tabletop.

\subsection{Additional Dataset Processing Details}\label{app:dataset}
Out of non-excluded objects, we ranked the quality of the provided data in order of (i) Google 16k mesh, (ii) Poisson reconstruction, and (iii) TSDF file. For instance, if the 16k mesh was available, we would always prefer to use that as the initial mesh for processing before the Poisson reconstructed mesh. The excluded objects and the exact reasons for their exclusion are listed in Table \ref{table:ycb_exclusions}.

\begin{table}[ht]
    \centering
    \begin{tabular}{|c|c|}
    \hline
    Object & Reason for Exclusion \\
    \hline
    \verb|019_pitcher_base| & too big \\
    \verb|022_windex_bottle| & poor model: transparency \\
    \verb|023_wine_glass| & poor model: transparency \\
    \verb|024_bowl| & thin walls \\
    \verb|025_mug| & thin walls \\
    \verb|026_sponge| & deformable, too flat \\
    \verb|028_skillet_lid| & poor model: transparency \\
    \verb|029_plate| & thin walls \\
    \verb|030_fork| & too flat \\
    \verb|031_spoon| & too flat \\
    \verb|032_knife| & too flat \\
    \verb|033_spatula| & too big \\
    \verb|035_power_drill| & too big \\
    \verb|037_scissors| & too flat \\
    \verb|038_padlock| & no file \\
    \verb|039_key| & no file \\
    \verb|040_large_marker| & too small \\
    \verb|041_small_marker| & too small \\
    \verb|042_adjustable_wrench| & too flat \\
    \verb|046_plastic_bolt| & no file \\
    \verb|047_plastic_nut| & no file \\
    \verb|049_small_clamp| & too small \\
    \verb|050_medium_clamp| & too small \\
    \verb|053_mini_soccer_ball| & too big \\
    \verb|059_chain| & multibody \\
    \verb|076_timer| & lost features, not interesting \\
    \hline
    \end{tabular}
    \caption{\textbf{YCB Exclusions.}}
    \label{table:ycb_exclusions}
    \vspace{-0.5cm}
\end{table}
While flat utensils are generally too flat to be picked up by an Allegro hand from a flat table, we did replace the excluded mug with a teacup from the ShapenetSem dataset \citeapp{savva2015_shapenetsem} with ID \verb|23fb2a2231263e261a9ac99425d3b306| and scaled by a factor of \verb|0.00038748778493825193|. This cup was added to the adversarial category.

\newpage
\subsection{Distribution of Normalized Min-Weight Values}\label{app:hist}
The histogram of values for the $\bar{\ell}^*$ values corresponding to successful and failed grasps optimized using FRoGGeR is shown here.
\begin{center}
    \includegraphics[width=\linewidth]{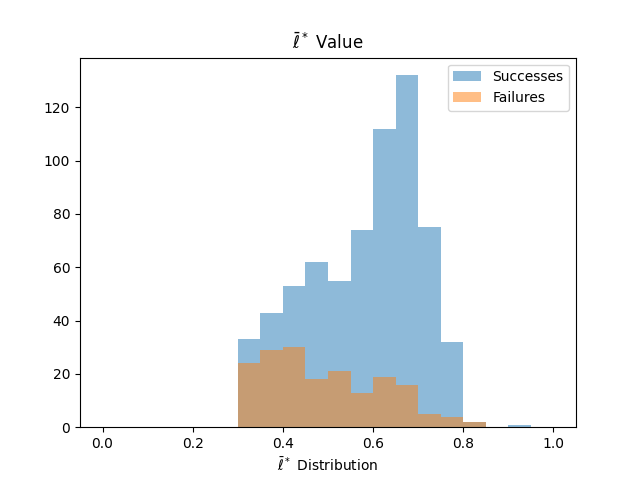}
\end{center}

\subsection{Other Experimental Parameters}\label{app:params}
For all experiments, we used a 4-sided pyramidal approximation of the friction cone. We enforced a minimum safety margin of $1mm$ between every collision geometry pair that was not a fingertip/object pair. For fingertip/object pairs, we allowed interpenetration up to $3mm$.

We selected a specific desired point of contact on each fingertip such that the forward kinematics were fixed. This point was located on each fingertip at an angle of $60^\circ$ tilted towards the palm, measured from the very tip of each finger. 

We supplied the following constraint tolerances to the optimization solver:
\begin{table}[ht]
    \centering
    \begin{tabular}{|c|c|}
    \hline
    Constraint & Tolerance \\
    \hline
    joint & 1e-2 \\
    surface contact & 5e-4 \\
    collision & 1e-3 \\
    force closure & 1e-5 \\
    \hline
    \end{tabular}
    \caption{\textbf{Constraint Tolerances.}}
    \label{table:constraint_tolerances}
    \vspace{-0.5cm}
\end{table}

We note that the force closure constraint refers to the robustness constraint for our method and the QP equality constraint for the baseline. In the original implementation of the method of \cite{wu2022_learningdexgraspsgenmodel} (obtained through private correspondence), the authors used a tolerance of 1e-7. In practice, we had to loosen this slightly to obtain a reasonable feasibility rate for their method. 

Finally, we use the default rigid body contact model implemented in \verb|Drake| for all of our simulations, the details of which we defer to the software documentation \cite{drake}.

\subsection{Acknowledgments}\label{app:acknowledgments}
We thank Victor Dorobantu for useful discussions involving our proposed Hessian approximation. We thank Ivan D. J. Rodriguez for help with setting up experiments. We thank Wu \textit{et al.} for their thoughtful correspondence concerning their work. We thank Philipp Wu for constructive feedback and comments. Finally, we thank all developers and maintainers of the open-source software that made this work possible (not cited in the main text but used either directly or indirectly: \citeapp{jax2018_github, trimesh, quantecon, lam2015_numba, pan2012_fcl}).

\bibliographystyleapp{unsrt}
\bibliographyapp{references}

\end{document}